\documentclass[twocolumn,conference]{IEEEtran}
\usepackage[latin9]{inputenc}
\usepackage{verbatim}
\usepackage{amsmath}
\usepackage{amssymb}
\usepackage{graphicx}
\usepackage[unicode=true,
 bookmarks=true,bookmarksnumbered=true,bookmarksopen=true,bookmarksopenlevel=1,
 breaklinks=false,pdfborder={0 0 0},pdfborderstyle={},backref=false,colorlinks=false]
 {hyperref}
\hypersetup{pdftitle={Your Title},
 pdfauthor={Your Name},
 pdfpagelayout=OneColumn, pdfnewwindow=true, pdfstartview=XYZ, plainpages=false}

\makeatletter
\usepackage[caption=false,font=footnotesize]{subfig}

\@ifundefined{showcaptionsetup}{}{%
 \PassOptionsToPackage{caption=false}{subfig}}
\usepackage{subfig}
\makeatother

\begin{document}

\title{Dynamic Control of Pneumatic Muscle Actuators}

\author{Isuru~S.~Godage$^{*}$, Yue~Chen$^{\dagger}$, and Ian~D.~Walker$^{\ddagger}$ }
\maketitle
\begin{abstract}
Pneumatic muscle actuators (PMA) are easy-to-fabricate, lightweight,
compliant, and have high power-to-weight ratio, thus making them the
ideal actuation choice for many soft and continuum robots. But so
far, limited work has been carried out in dynamic control of PMAs.
One reason is that PMAs are highly hysteretic. Coupled with their
high compliance and response lag, PMAs are challenging to control,
particularly when subjected to external loads. The hysteresis models
proposed to-date rely on many physical and mechanical parameters that
are difficult to measure reliably and therefore of limited use for
implementing dynamic control. In this work, we employ a Bouc-Wen hysteresis
modeling approach to account for the hysteresis of PMAs and use the
model for implementing dynamic control. The controller is then compared
to PID feedback control for a number of dynamic position tracking
tests. The dynamic control based on the Bouc-Wen hysteresis model
shows significantly better tracking performance. This work lays the
foundation towards implementing dynamic control for PMA-powered high
degrees of freedom soft and continuum robots.
\end{abstract}

\section{Introduction\label{sec:Introduction}}

\begin{table}[b]
$*$ School of Computing, DePaul University, Chicago, IL 60604. email:
\href{mailto:igodage@depaul.edu}{igodage@depaul.edu}. $\dagger$
Dept. of Mechanical Engineering, University of Arkansas, Fayetteville,
AR 72701. $\ddagger$ Dept. of Electrical and Computer Engineering,
Clemson University, SC 29634. \vspace{2mm} \\ This work is supported
in part by the National Science Foundation grant IIS-1718755.
\end{table}
Pneumatic muscle actuators (PMA) are a popular choice for powering
soft and continuum robots \cite{daerden2002pneumatic}. There lightweight
design, high compliance and high power-to-weight ratio, combined with
the ease of fabrication and customization have fueled their popularity
among the researchers and hobbyists alike \cite{daerden2002pneumatic}.
Invented in late 50's (also known as Mckibben artificial muscles \cite{tondu2000modeling})
PMAs have been well studied over the years and commercialized for
industrial applications \cite{musclefesto}. Based on the same fundamental
operation principle, researchers have investigated novel varieties
of PMA actuators to generate nonlinear and complex deformations beyond
the linear (extending or contracting) strain of traditional PMAs \cite{krishnan2015kinematics}.
Moreover, the PMA's also laid the foundation for novel types of fluidic
muscle actuators, such as fiber-reinforced soft bending actuators
\cite{polygerinos2015modeling}, now widespread in soft robotics \cite{tolley2014resilient}.

Unlike the soft bending actuators \cite{polygerinos2015modeling},
PMA powered robots, such as multisection continuum arms \cite{godage2011shape},
can operate at much higher pressure levels, and therefore are able
to generate higher forces to execute useful tasks in the task-space.
The stiffness of PMAs, which is a function of the pressure provided,
could be varied within a wider range to attain compliance for environmental
interactions and stiffness for supporting body weight during manipulation
\cite{walker2005continuum} and locomotion \cite{godage2012locomotion}.
For instance, the well known OctArm continuum robotic manipulator,
developed at Clemson University by Dr. Walker and the group demonstrated
a range of applications including compliant manipulation of fragile
objects as well as manipulating and dragging heavy objects \cite{mcmahan2006field}. 

The recent surge in soft robotics and compliant human-friendly robotics
have collectively put the spotlight back on compliant actuators such
as PMAs \cite{rus2015design}. Despite the widespread usage and research
conducted on soft and continuum robots, which has spanned over a decade
and half, PMA powered robots are still largely confined to laboratory
settings with their demonstrated potential untapped. This lag can
be attributed to the lack of effective dynamic control schemes developed
for such robots for handling the compliance and hysteresis; which
essentially leads to better PMA dynamic models. For instance, for
systems as complex as a traditional robot manipulator (with 7+ degrees
of freedom), the continuum robot state-of-the-art research lags in
terms of dynamic control and efficient dynamic models. The latter
however has seen significant advancement lately \cite{godage2016dynamics}. 

\begin{figure}[t]
\begin{centering}
\includegraphics[width=1\columnwidth,height=2.8cm]{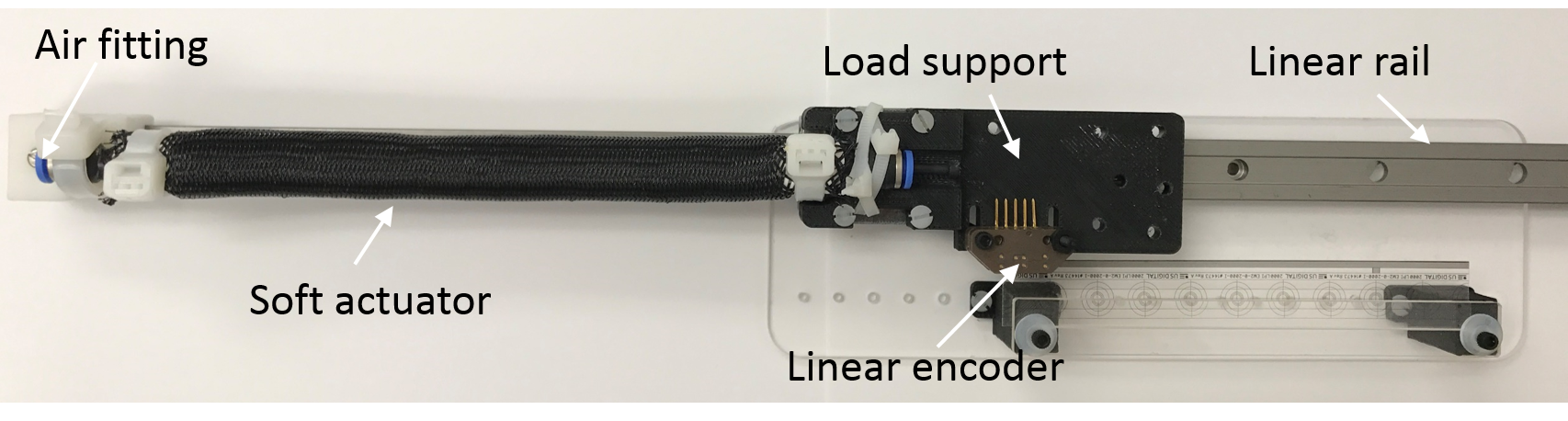}
\par\end{centering}
\caption{Pneumatic muscle actuator (PMA) experimental setup, detailing the
separate elements.}
\label{fig:exp_setup}
\end{figure}

The overall dynamics of continuum arms heavily depend on the dynamics
of PMAs. Yet, most of the dynamic models for continuum robots employs
intermediate joint-space variables such length change of PMAs while
pressure is being the true controlled variable. In quasi-static conditions,
one can consider that the length variation of a PMAs is proportional
to pressure. However, because of the hysteresis, this assumption does
not hold for dynamic motion, which is a requirement for robots to
efficiently operate in the task-space or match the human operation
bandwidth in human spaces as co-robots. 

Prior work have taken different avenues to model PMAs \cite{reynolds2003modeling,thanh2006nonlinear}.
Bulk of the work focuses on static or quasi-static models and propose
methodologies to systematically derive the models based on physical
parameters (such as bladder dimensions) and mechanical properties
(i.e., elastic coefficient etc.) of the construction material. However,
these properties are difficult to measure reliably, especially when
they are heavily coupled to one another during PMA development. Thus,
models have been limited to theoretical studies with inadequate experimental
testing, notably related to dynamics. Relatively low attention has
been paid for developing hysteresis models for PMAs \cite{thanh2006nonlinear}
and limited work has shown experimental evaluation under external
load \cite{chou1996measurement}. Also, tests carried out to-date
were also limited to tracking signals of bandwidth less than 0.5~Hz.
The authors introduced a variant of Bouc-Wen hysteresis model for
PMAs in \cite{godage2012pneumatic}. The method relies on experimental
characterization to identify the actuator specific dynamic behavior
and hysteresis. This approach is particularly suitable for accounting
for varying performance of PMAs due to complex and nonlinear interactions
of materials and variations of the fabrication method. Utilizing the
Bouc-Wen hysteresis model, the authors have shown that the model is
able to correctly simulate the hysteretic behavior of PMAs individually
as well as a continuum section (where three PMAs are bundled together
for generating spatial bending).

This work extends the contribution reported in \cite{godage2012pneumatic}
and presents the initial results on the implementation of dynamic
control for PMA's based on the Bouc-Wen hysteresis model. The paper
is organized as follows. Section \ref{sec:Materials-and-Methods}
details the experimental setup, the dynamic model and the details
of system identification process and the controller design. Section
\ref{sec:Experimental-Results} presents the experimental results
and compares the kinematic feedback control performance to the computed
torque control output that utilizes the Bouc-Wen hysteresis model
followed by the concluding remarks in Section \ref{sec:Conclusions}.

\section{Materials and Methods\label{sec:Materials-and-Methods}}

\subsection{Experimental Setup\label{subsec:Experimental-Setup}}

Figure \ref{fig:exp_setup} shows the experimental setup consisting
of the prototype PMA, high resolution (2000 quadrature counts per
inch) linear optical encoder, variable external load support, and
air supply to the PMA. The PMA construction is similar to the one
detailed in \cite{godage2012pneumatic}. The bladder is a silicone
tube of 12~mm diameter with 1.5~mm wall thickness, enclosed within
a 14~mm diameter Nylon braided mesh, and mounted on 4~mm pneumatic
union connectors at either end. The PMA has 170~mm unactuated length
($l_{0}$) and 85~mm steady state extension (calculated from 10 measurements
taken 100~s after applying the pressure step) at 0.4~MPa. The PMA
has a 0.022~kg mass where the moving carriage has 0.045~kg mass.
The PMA is attached between an immobile base and the low-friction
moving carriage (McMasterCarr part \# 6250K42), mounted on a linear
rail (McMasterCarr part \# 6250K3), to ensure that the PMA changes
length axially. The air pressure to the PMA is controlled by a digital
proportional pressure regulator (Pneumax 171E2N.T.D.0005S) that is
controlled via an analog, 0-10~V (maps to 0-0.9~MPa) voltage input
provided through a National Instruments PCI-6703 data acquisition
interface card. The quadrature encoder pulses are counted using a
CONTEC CNT-3208M-PE timer/counter interface card. The interface cards
are mounted on a Matlab Simulink Realtime target machine and controlled
directly from a Simulink model on a host computer and solved using
an ODE14X solver at 1~kHz. This high update rate ensures minimal
delay and accurate dynamic control performance.
\begin{center}
\begin{figure}[t]
\begin{centering}
\includegraphics[width=0.95\columnwidth,height=5cm]{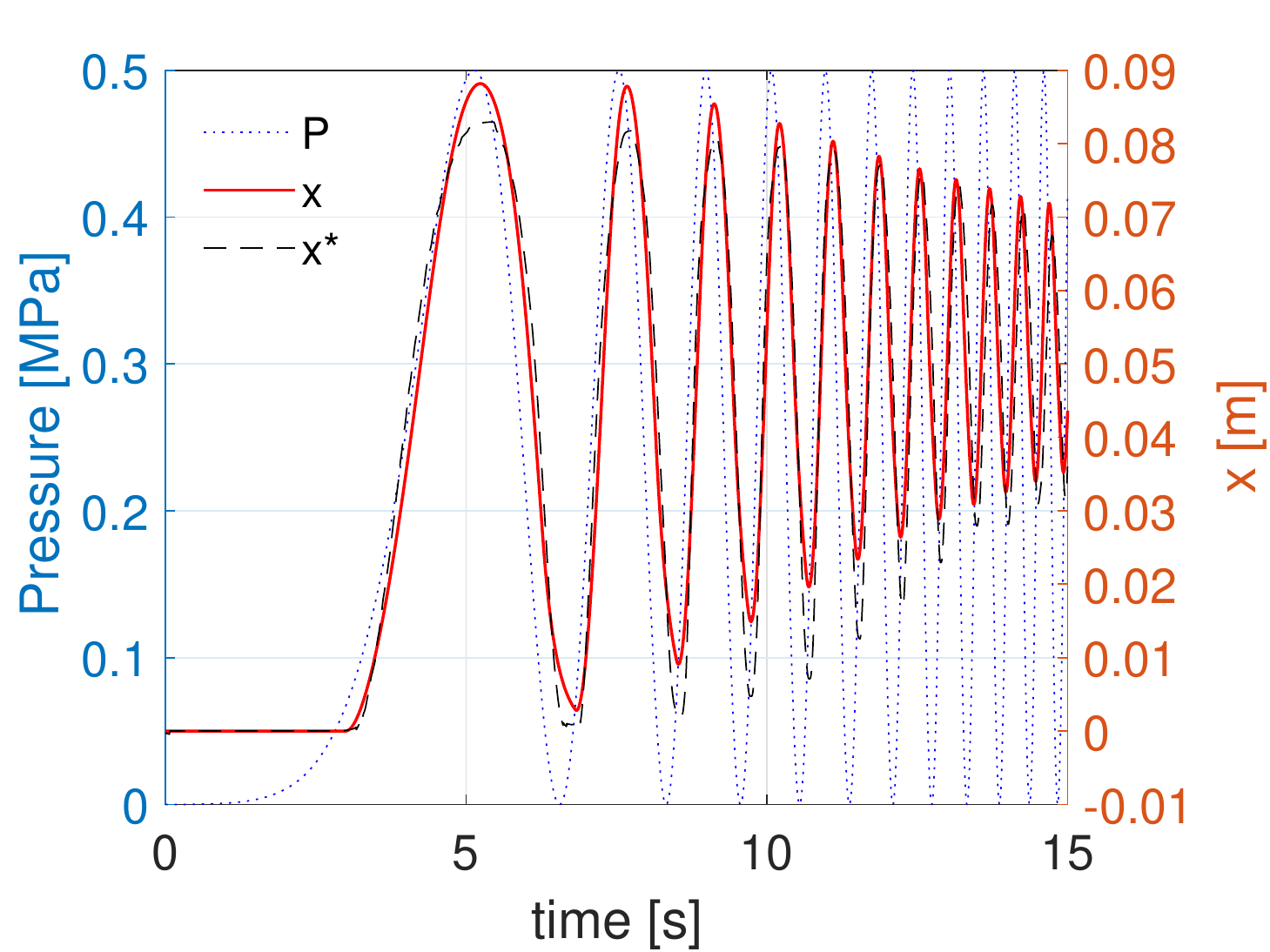}
\par\end{centering}
\caption{The plot of the sinusoidal chirp input pressure signal, system response
to the input signal, and the simulated system response with the use
of optimal Bouc-Wen shape coefficients.}
\label{fig:SystemResponseOptimal}
\end{figure}
\par\end{center}

\subsection{Dynamic Model with Hysteresis\label{subsec:Dynamic-Model-with}}

\begin{figure}[b]
\begin{centering}
\includegraphics[width=0.5\columnwidth]{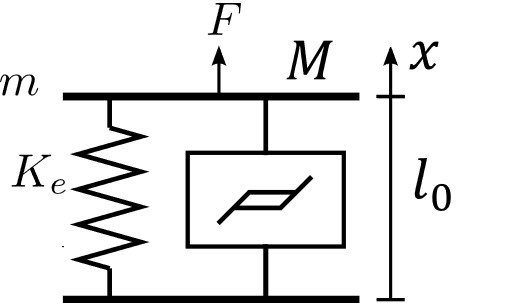}
\par\end{centering}
\caption{PMA model with the Bouc-Wen hysteresis block.}
\label{muscle_model}
\end{figure}

Schematic diagram of the dynamic system model we used in this work
is shown in Fig. \ref{muscle_model}. The carriage mass, $M$, to
which the free end of the PMA is considered as a load on the system.
Similar to \cite{godage2012pneumatic}, using the Lagrangian mechanical
principles, the PMA dynamic model can be derived as a single degree
of freedom system given by 

\begin{align}
\left(M+m\right)\ddot{x} & +(M+m)g+K_{e}x+z=Ap\label{eq:dynaModel}\\
\dot{z} & =\dot{x}[\alpha-\{\beta\text{sgn}\left(\dot{x}z\right)+\gamma\}|z|]\label{eq:Bouc-wenForce}
\end{align}
where $M$, $m$, $K_{e}$, $g$, and $x$ are the variable load mass,
PMA mass, PMA bladder linear elastic stiffness, gravitational acceleration,
and PMA length change. For an in-depth treatment of the equation of
motion (EoM) derivation, the readers are referred to \cite{godage2012pneumatic}.
$\alpha>0\in\mathbb{R}$, $\beta>0\in\mathbb{R}$, and $\gamma\in\mathbb{R}$
are dimensionless Bouc-Wen hysteresis loop control parameters. $A$
is the cross-section are of the PMA and $p$ is the supplied pressure
(joint-space variable).

The shape of the hysteresis curve can be matched to that of the experimental
results by finding the appropriate values for $\alpha$, $\beta$,
and $\gamma$. To realize this, we apply the PMA with a signal of
varying frequency to ensure to capture as much dynamic response information
as possible in order to model the system well in faster motion. The
PMA was provided with a 0.5~MPa sinusoidal chirp signal of frequencies
from 0.1~Hz to 3~Hz for a 15~s duration and the input pressure
signal and the system response, $x'$, were recorded. This experiment
is repeated 10 times and the average system response is computed by
taking the mean response at each time step. The frequency range and
the variation ensure that both quasi-static (low frequency), transient,
and dynamic behavior of the system are captured. The experimentally
recorded data $\left(x'\right)$, are then used to identify the Bouc-Wen
hysteresis shape coefficients. To achieve this, we implemented the
EoM, given by \eqref{eq:dynaModel} and \eqref{eq:Bouc-wenForce},
in Matlab Simulink and defined the cost function, $c\in\mathbb{R}$,
given by

\begin{align}
c\left(\alpha,\beta,\gamma,d,K_{e},p_{dz}\right) & =\text{RMS}\sum_{\forall t}\left(x-x'\right)\label{eq:costFn}
\end{align}
where $t$ denotes a specific time instance when we measured the experimental
data and $x$ is the simulated system response. To derive the cost
function for the entire simulation, we compute the vector of differences
of the simulated and experimental data for each time sample and then
take the root mean square (RMS) value of this array. In addition to
hysteresis shape parameters, we include the system damping $\left(d\right)$,
elastic coefficient $\left(K_{e}\right)$, and the PMA dead zone $\left(p_{dz}\right)$
as parameters to be modeled for the setup shown in Fig. \ref{fig:exp_setup}.
The reason for including the latter parameters is that it is challenging
to experimentally measure them reliably in dynamic motion.

We then employed the Matlab global search functionality and ran a
constrained optimization routine (using 'fmincon') until we find the
optimal Bouc-Wen hysteresis shape parameters. Figure \ref{fig:SystemResponseOptimal}
also shows the simulated response, obtained from \eqref{eq:dynaModel},
using the optimal Bouc-Wen shape parameters given by $\alpha=23.705$,
$\beta=1.7267$, and $\gamma=-42.593$ where as $d=155.76$, $K_{e}=624.78$,
and $p_{dz}=66.922\,\text{kPa}$. The optimized numerical model's
output is then plotted alongside the experimental data.  It can be
seen that the numerical model captures the system dynamics, both steady
state and dynamics, well overall.%

\section{Experimental Results\label{sec:Experimental-Results}}
\begin{center}
\begin{figure}[b]
\begin{centering}
\includegraphics[width=0.7\columnwidth]{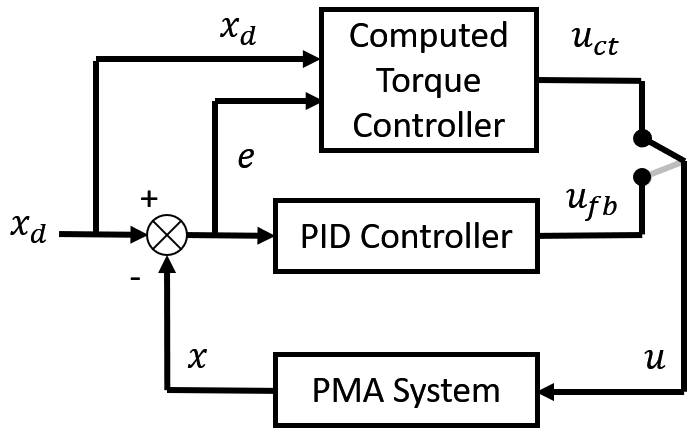}
\par\end{centering}
\caption{Computed torque controller and the standard joint-space PID controller.}
\label{fig:controller}
\end{figure}
\par\end{center}

The dynamic model is then used to implement the control system shown
in Fig. \ref{fig:controller} which includes the standard joint-space
feedback (kinematic) PID controller and the computed torque controller
\cite{murray2017mathematical} utilizing the dynamic model described
in Section \ref{subsec:Dynamic-Model-with}. The controller is implemented
in Matlab Simulink Realtime and switched between the two for assessing
the control performance. In the Simulink Realtime model, the inner
control loop was run at 100~Hz where the input signals to the pressure
regulator was run at 20~Hz. This low outer control loop ensures that
the digital pressure regulator bandwidth is matched without driving
the inner pressure control loop unstable. As the load, we use a 500~g
weight and we erected the experimental setup so that the gravity induced
weight is toward the PMA extension (downwards). This prevents the
PMA from buckling under the load as here, there is no constraining
mechanism like the one reported in \cite{godage2012pneumatic}. 

Then we applied the sinusoidal position tracking signal given by

\begin{align}
x_{d} & =0.005+0.0225\sin\left(2\pi ft\right)\label{eq:trackingSignal}
\end{align}
where $f$ is the frequency of the signal. Note the 0.005~m bias
we applied to the tracking signal to compensate initial position offset
caused by the applied external load.

The performance of PID feedback control and the proposed computed
torque controller that uses the Bouc-Wen hysteresis model was then
evaluated. The sub-figures under each test shown in Fig. \ref{fig:expResults}
compare the position tracking error for both PID control and computed
torque controllers. %
Figure~\ref{fig:expResults}-a shows the controller performance results
for $f=0.5~\text{Hz}$ signal tracking. Note that 0.5~Hz is relatively
fast signals to track in comparison to prior work on PMA control which
typically include frequencies less than 0.5~Hz. The computed torque
controller performance is superior to PID controller particularly
in phase response. This is significant, especially when comparing
the inherent lag present in PMAs. For instance, the system response
for the characterization signal shown in Fig.~\ref{fig:SystemResponseOptimal},
exhibit this phase lag of the output. This lag is reflected on the
PID controller output and becomes a significant factor when tracking
high frequency signals such as 1~Hz (Fig.~\ref{fig:expResults}-b).
This system delay, to a lesser degree, also affects the computed torque
controller but yet shows better tracking performance where PID controller
fails to maintain the pace of tracking. Note that the dynamic controlled
response exhibits about 10\% overshoot but quickly corrects quickly
to follow the falling edge accurately. We then pushed our controllers
to track a 2~Hz signal and the results are compared in Fig.~\ref{fig:expResults}-c.
Here, both controllers were unable to track the signal well, although
the computed torque controller output was better overall and within
an acceptable, less than $180^{\circ}$, phase lag and overall tracking
amplitude profile. Whereas the PID control output was $180^{\circ}$
out of phase. Hence, the proposed work demonstrates strong potential
to be useful in realizing PMA-powered multisection continuum arm dynamic
control towards applications in human spaces.
\begin{center}
\begin{figure}[tbh]
\begin{centering}
\subfloat[]{\begin{centering}
\includegraphics[width=1\columnwidth]{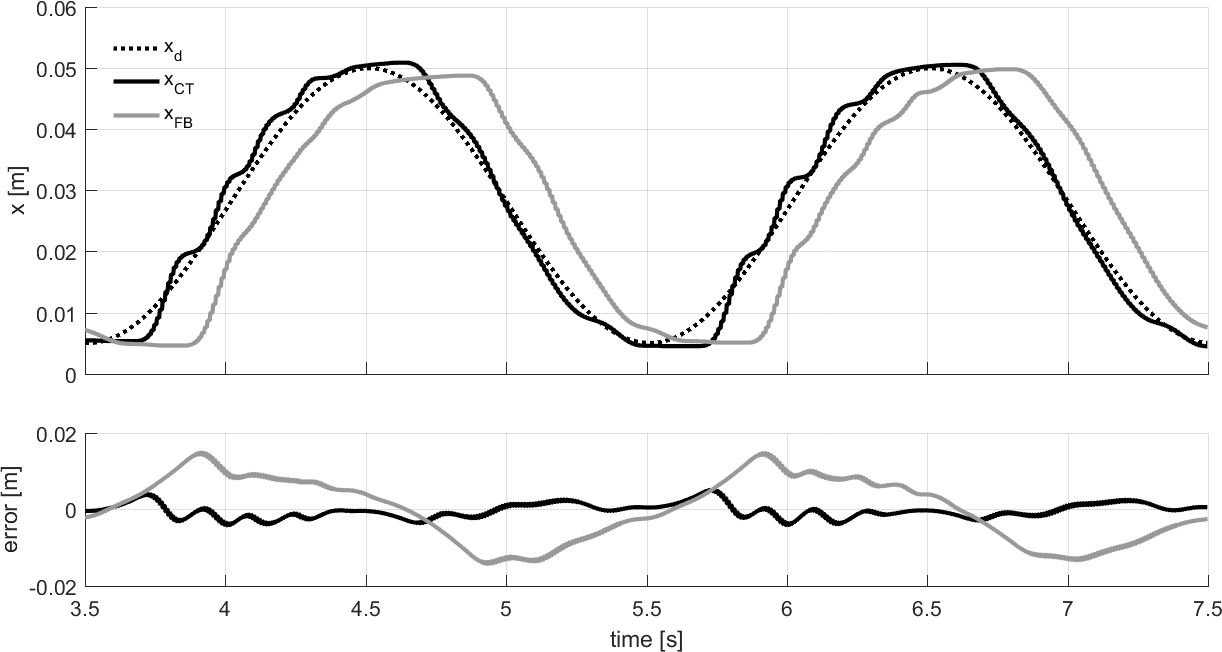}
\par\end{centering}
}
\par\end{centering}
\begin{centering}
\subfloat[]{\begin{centering}
\includegraphics[width=1\columnwidth]{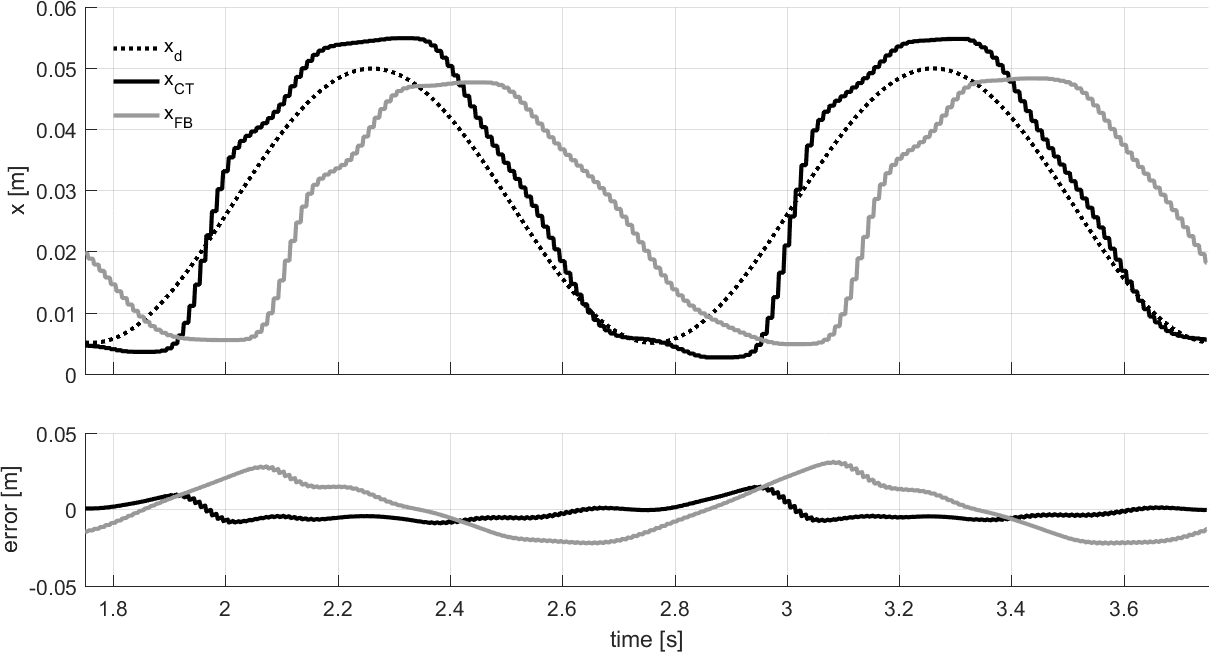}
\par\end{centering}
}
\par\end{centering}
\begin{centering}
\subfloat[]{\begin{centering}
\includegraphics[width=1\columnwidth]{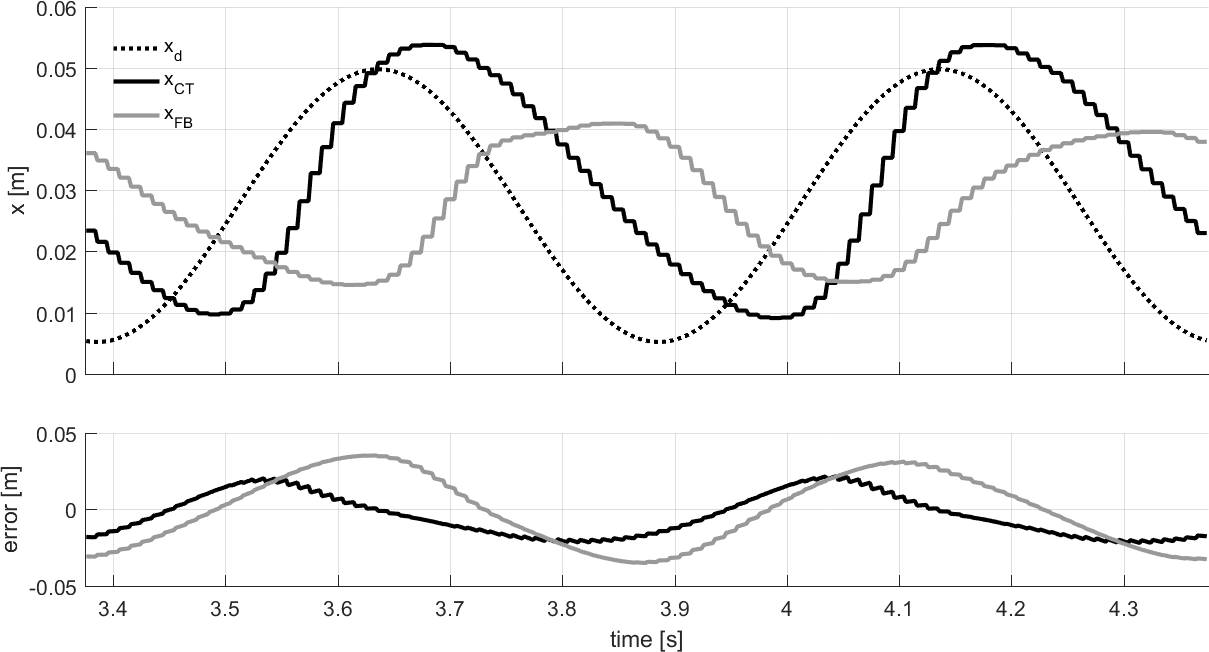}
\par\end{centering}
}
\par\end{centering}
\caption{Computed torque controller and the standard joint-space PID controller
performance comparison, (a) 0.5~Hz frequency, (b) 1.0~Hz, and (c)
2.0~Hz. The plots below of each experiments shows the tracking error.
$x_{d}$ is the desired tracking signal, $x_{FB}$ is the PID controller
output, and $x_{CT}$ is the computed torque controller output.}
\label{fig:expResults}
\end{figure}
\par\end{center}

\section{Conclusions and Future Work\label{sec:Conclusions}}

Pneumatic muscle actuators (PMAs) are the choice of many, researchers
and hobbyists alike for powering soft and continuum robots. PMA desirable
features include inherent compliance and high power-to-weight ratio.
PMA operation is highly hysteretic, posing challenges in controlling
them, particularly in highly dynamic applications. Consequently, the
soft and continuum robots so far have been limited to teleoperation
(open-loop control) or kinematic (slow motions) control thus limiting
their application potential. We proposed a dynamic control scheme
for PMAs, which is based on the dynamic model previously proposed
by the authors, where hysteresis is modeled via a Bouc-Wen hysteresis
model. The work compared the dynamic control performance to a PID
feedback controller for a number of dynamic position tracking tests
on a PMA with external loading. The results showed that the proposed
dynamic controller is capable of tracking the dynamic signals better.
We plan to extend the proposed dynamic control approach to the continuum
arm reported in \cite{godage2016dynamics} in manipulation tasks in
human spaces.

\bibliographystyle{IEEEtran}
\bibliography{Biblio}

\end{document}